\documentclass[letterpaper]{article}
\usepackage{aaai}
\usepackage{times}
\usepackage{helvet}
\usepackage{courier}
\usepackage{xcolor}
\usepackage{siunitx}
\usepackage{graphicx}
\usepackage{amsfonts}
\usepackage{amsmath}
\usepackage{multirow}
\usepackage{float}

\DeclareMathOperator*{\argmin}{arg\,min}

\begin{document}
%
\title{Transfer learning to improve streamflow forecasts in data sparse regions}
\author{Roland Oruche, Lisa Egede, Tracy Baker, Fearghal O'Donncha\\
Department of Electrical Engineering \& Computer Science, University of Missouri–Columbia, USA\\
Human Computer Interaction Institute, Carnegie Mellon University, USA\\
The Nature Conservancy, New York, NY, USA \\
IBM Research Europe, IE\\
rro2q2@umsystem.edu, legede@cs.cmu.edu, tracy.baker@tnc.org, feardonn@ie.ibm.com 
}
\maketitle
\begin{abstract}
Effective water resource management requires information on water availability – both in terms of quality and quantity - spatially and temporally. 
In this paper, we study the methodology behind Transfer Learning (TL) through fine-tuning and parameter transferring for better generalization performance of streamflow prediction in data-sparse regions. We propose a standard recurrent neural network in the form of Long Short-Term Memory (LSTM) to fit on a sufficiently large source domain dataset and repurpose the learned weights to a significantly smaller, yet similar target domain datasets. We present a methodology to implement transfer learning approaches for spatiotemporal applications by separating the spatial and temporal components of the model and training the model to generalize based on categorical datasets representing spatial variability. 
The framework is developed on a rich benchmark dataset from the US and evaluated on a smaller dataset collected by The Nature Conservancy in Kenya.
The LSTM model exhibits generalization performance through our TL technique. 
Results from this current experiment demonstrate the effective predictive skill of forecasting streamflow responses when knowledge transferring and static descriptors are used to improve hydrologic model generalization in data-sparse regions.
\end{abstract}


\section{Introduction}

Machine learning has had limited impact in hydrological applications due primarily to data sparsity, the large spatial and temporal extents involved, and the challenges of generalising to conditions not observed in the training data (particularly critical in the face of changing climate). Instead, practitioners rely on complex hydrological modeling tools that are cumbersome to set up, and generally require an expert both to configure and interpret results. While some studies have explored LSTM networks to forecast hydrological conditions, these have been applied on benchmark datasets from regions with well developed, long-term environmental monitoring programs \cite{addor2017camels}. In this paper, we propose a framework based on transfer learning and categorical descriptors of pertinent hydrological and geographical features to forecast streamflow discharge in the Upper Tana region in Kenya. 

\begin{figure}[t!]
  \centering
  \includegraphics[width=\linewidth]{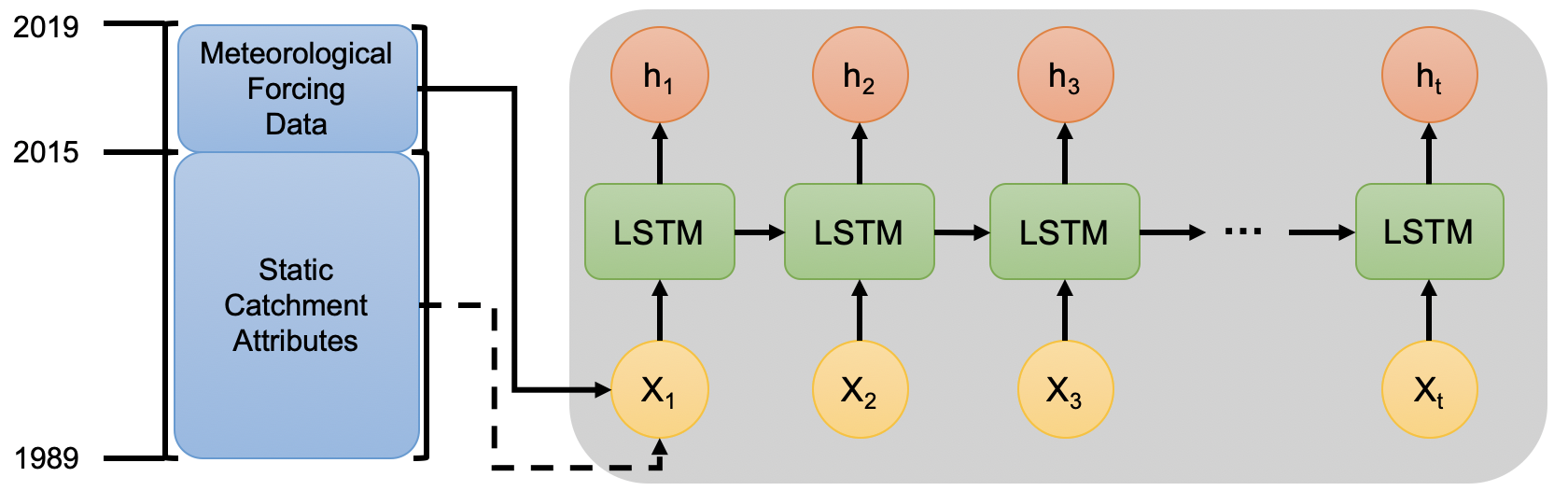}
  	  \caption{LSTM static descriptors architecture}
  \label{fig:lstm-static-arch}
\end{figure}

The 2018 edition of the United Nations World Water Development Report stated that nearly 6 billion people will suffer from clean water scarcity by 2050 with some suggestions that even this number is an underestimate \cite{boretti2019reassessing}. Climate change, population growth, and pollution are depleting water sources across the globe, threatening the livelihoods of millions of people. 
The Nature Conservancy (TNC) is a globally active conservation organization whose mission is to conserve the water and lands on which life depends and to support disparate water restoration projects. A key tool in the TNC toolbox is \textit{Water Funds}, which aims to support water conservation and restoration activities by investing in land management to improve water quality and quantity and generate long-term benefits for people and nature. 

TNC has created 35 Water Funds in 12 countries and they project that by 2025, water funds can be capable of providing 70 million people with water security, improving livelihoods for 150,000 rural community members, and protecting 2 million hectares of freshwater in Latin America, Asia, and Africa \cite{TNC2020}. 
A fundamental challenge to water fund success is choosing the most effective soil and water conservation and land management strategies while also remaining cognisant of the local environment and human activities. 
Another critical challenge is that there is often little historical data available for accurately representing the rainfall-runoff relationship and it is often  infeasible to collect sufficient (long-term) observation data to characterise water quality and quantity over large regions. For example, the Upper Tana Water Fund in Kenya covers an area approximately \SI{17,000}{km^2}), and data is not well distributed temporally and spatially. 
It is common practice to then fit physically based models to the extent possible to describe the system's potential response to changes. 

The most widely used hydrological models to use in data scarce regions is the Soil and Water Assessment Tool (SWAT), which has seen widespread usage in Africa \cite{akoko2021review} and globally. SWAT is a  continuous‐time model that can operate on daily, monthly, and annual time steps and was designed to predict the potential impact
of land management practices on water quality and quantity in large ungauged agricultural watersheds. The model is both physically and empirically  based,  computationally  efficient,  and capable of continuous
simulation over long periods. SWAT was developed by the USDA Agricultural Research Services with development beginning in 1980 with many additional modules implemented in the intervening years \cite{gassman2007soil}. Major model components include  weather, hydrology, soil temperature and properties, plant  growth,  nutrients,  pesticides,  bacteria  and  pathogens, and land management. The model has been applied numerous times in the Tana River Basin \cite{jacobs2007mitigating,hunink2011impacts,njogu2017assessment,vogl2017valuing}.

The use of machine learning in hydrology has shown promising results for tasks such as flood forecasting and run-off prediction \cite{young2017physically,kratzert2019neuralhydrology,nearing2020deep,kratzert2019toward}. However, these successes typically occur in countries with well developed environmental monitoring programs and relatively plentiful observation data to train on. No previous studies have demonstrated the ability to effectively generalize to regions where data are scarce and inconsistent due to the lack of temporal and geospatial resources (geographical constraints, lack of population, funding sources).

In this paper, we applied Transfer Learning (TL), in which the knowledge manifested in one task is applied (or transferred) to similar objective tasks. Specifically, we trained a Long Short-Term Memory (LSTM) on a source domain, and fine-tuned the resulting model towards a target domain. We refer to the \textit{source} as the collection of inputs used to train the LSTM model, while the \textit{target} is the significantly smaller dataset used to fit and evaluate the model's performance. We employed a transfer learning technique that optimizes the weights during training and fine-tunes the parameters on a selected basin during validation. These parameters were then transferred to a replicated multi-layered LSTM network to generalize the streamflow response over different regions. We trained and optimized our model on our source domain using the US Catchment Attributes and Meteorology for Large-Sample Studies (CAMELS-US) dataset, a benchmark dataset that covers 531 basins across the US region \cite{newman2014development,newman2015development}. We then performed TL over our target domain to generalize the performance using a dataset we curated with The Nature Conservancy (TNC) which covers the Upper Tana River Basin in Kenya. Previous studies have demonstrated that combining categorical geospatial features (e.g., land cover, climate metrics, soil properties) in addition to corresponding dynamic inputs (precipitation, air temperature) helps improve the performance of streamflow prediction for ML-based hydrological models \cite{kratzert2019toward}. Influenced by this, the structure of the Kenya dataset generated the same static basin feature as additional input to our LSTM model as illustrated in Figure \ref{fig:lstm-static-arch}. We validated the utility of this approach by conducting extensive experiments to demonstrate effective predictive skill of a standard LSTM model with transfer learning and static descriptors.

The contributions of this paper are as follows:
\begin{itemize}
    \item We present a novel dataset of hydrological and categorical catchment attributes collected from the Upper Tana River Basin in Kenya, which follows the same structure as the benchmark CAMELS dataset. We highlight examples of scarcity in the temporal and geospatial data collected.
    
    \item We propose a transfer learning approach that optimizes the models performance over the CAMEL-US source domain dataset to address the issues of data sparsity in smaller datasets. We use a fine-tuning method that randomly selects a basin that's performance score was less than 50\% the median distribution. The model's parameters are then transferred over the TNC dataset to fit and predict the streamflow response.
    
    \item We present experimental results that compare models using transfer learning and static descriptors to its counterparts. These extensive experiments show the model's effective predictive skill of forecasting streamflow response where there lacks sufficient data needed for effective generalization.
    
    \item Finally, we discuss further research opportunities to apply machine learning to improve water management and conservation practises.
\end{itemize}

\section{Related Work}
Recent advancements in machine learning has led to widespread interest amongst hydrologists and environmental scientists as a solution to address the challenges that persist with streamflow and run-off forecasting. While previous works have approached performance levels of state-of-the-art physics-based methods \cite{hsu1995artificial,kratzert2019toward,nearing2020deep}, the challenge remains whether it can generalize in different domains, and specifically if it can perform in regions with limited training data. Hence, we review the existing literature on transfer learning techniques and its applicability for use cases around hydrology.

Previous works such as \cite{pan2009survey} categorize the types of transfer learning as \textit{inductive}, \textit{transductive}, and \textit{unsupervised}, which has built foundation for many techniques used in machine learning, more specifically, time-series forecasting \cite{howard2018universal,fawaz2018transfer,ye2018novel}. For instance, Lin et al. \cite{lin2018transfer} forecasts traffic conditions in poorly monitored urban areas through a transfer learning approach that extracts useful spatio-temporal features from data-rich traffic areas. McCarthy et al. \cite{mccarthy2019amsterdam} has shown that using a unified vector representation approach between the source and target data to transfer knowledge in heterogeneous feature spaces improves flight delay forecasting. Fawaz et al. \cite{fawaz2018transfer} uses convolutional neural networks (CNNs) to invoke transfer learning by fine-tuning the weights of a pre-trained CNN model over the UCR time-series benchmark dataset for classification tasks. In cases relevant to hydrology, Kimura et al. \cite{kimura2020convolutional} implements a feature-representation transfer approach via a CNN from time-series data for predicting floods. The CNN takes advantage of low-level feature extraction to transfer knowledge in a target domain based on a pre-processing conversion mechanism between the spatial imagery and time-dependent data inputs. Gang et al. \cite{gang2021improving} addresses the data-sparsity issue in small and inconsistent target datasets via transfer learning to improve the performance of flood forecasting in both data-rich and data-sparse urban areas. The success of transfer learning, however, has mainly been applied in computer vision for low to mid-level feature mapping or implementing a freezing mechanism on CNNs to reduce computational overhead. There presents a lack of literature on transfer learning used for LSTM networks and other variations of RNNs. In addition, data sparsity and inconsistency remains a critical issue behind the performance degradation in generalization. 

Physics- or empirical-based hydrological models are well established in the literature with research in the space receiving significant impetus with the US Clean Water Act of 1977. In these systems, the hydrological processes of water movement are represented by finite difference equations that are resolved over defined spatial and temporal ranges. Data inputs to resolve streamflow processes include meteorological forcing and a large number of  parameters  describing  the  physical  characteristics  of  the  catchment (soil moisture content, initial water depth, topography, topology, dimensions of river network, etc.) \cite{devia2015review}. Popular modelling systems include SWAT \cite{arnold2012swat}, MIKE SHE \cite{graham2005flexible}, and the VIC framework \cite{gao2010water}. On the SWAT model alone, there are over 4,500 peer-reviewed journal articles describing its application to different hydrology studies \cite{swat_lit_db_2021}.

More recently, extensive research efforts have focused on the potential of deep learning (DL) for hydrology studies \cite{shen2018transdisciplinary,shamshirband2020predicting}. In particular, research has focused on the potential of recurrent networks and LSTM to resolve the complex, nonlinear, spatiotemporal relationship between meteorological forcing, soil moisture and streamflow \cite{kratzert2019toward}. In a provocative recent paper, \cite{nearing2021role} argued that there is significantly more information in large-scale hydrological data sets than hydrologists have been able to translate into theory or models. This argument for increased scientific insight and performance from machine learning rests on the assumption that large-scale data sets are available globally (over sufficient historical periods) to condition and inform on hydrological response. While significant progress on hydrology dataset curation has been achieved in the US \cite{newman2014development,newman2015development}, and Europe \cite{klingler2021lamah} this is not implemented for many other regions. Inspired by the existing research contributions, we investigate and apply a transfer learning approach to address the issue of data sparsity when forecasting the streamflow response of catchment areas. 


\section{Data Collection}
The Tana River supplies 95\% of Nairobi’s water and half of the country’s energy. The entire Tana River Basin covers $\sim$\SI{95,000}{km^2} and is home to millions of wildlife species as well local communities \cite{TNC2015}. Since the 1970s, the number of small subsistence farms in the upper basin has skyrocketed \cite{TNC2015}. The dual effects of climate change and rapid population growth are leading to increased sediment erosion in the Upper Tana River Basin, reducing the capacity of reservoirs, and increasing water treatment costs. Water
scarcity is predicted to disproportionately affect communities that have least contributed to
climate change, such as the millions of farmers and fishermen who rely on the Tana River and its
tributaries for survival and economic prosperity. Taking action now to prevent further damage and help these communities preserve their water resources is critical to ensure the region’s water security for the future. To this end, The Nature Conservancy created a water fund, an
organization that brings together public and private stakeholders to address threats to water
security at their source through targeted, long-term investments in watershed conservation and management activities \cite{TNC2015}. In 2015, The Nature Conservancy established the Upper Tana-Nairobi Water Fund as the first of its kind in Africa. 
The Water Fund aims to improve water quality and quantity for all stakeholders, and is founded on the principle that it is less expensive to prevent water problems at the source than it is to address them further downstream. 

The primary focus of the Upper Tana-Nairobi Water Fund (UTNWF) is to improve farming practices in the watershed. Water Fund partners work with almost 50,000 farmers to provide the skills, training and resources they need to conserve water, reduce soil runoff, and improve productivity. Effective management of water conservation activities require quantitative metrics on differences in water availability in different areas of the watershed, subject to different water conservation strategies (and control watersheds with no intervention). 
Previously, TNC have relied on a combination of SWAT model simulations together with data measurements to inform Water Fund operations \cite{baker2013using}. However, this comes with the challenges outlined earlier related to computational cost, complexity, and need for high level of user expertise.

UTNWF collects a relatively rich hydrologic dataset that provides opportunities for DL-based forecasting. Data were collected from February 2015 to January 2020 at 26 stream locations as described in Table 1 in \cite{Leisher2018}. Data on water level and temperature were collected at 30-minute intervals using Onset HOBO water level data logger. HOBO loggers are a robust, widely-used tool to continuously monitor water level and temperature with specified device accuracy of 0.1\%. A standard challenge in hydrology is that water level data are easy to measure with low cost equipment while flow or discharge are difficult to measure requiring sophisticated equipment and experienced personnel. On the other hand, water level communicate little meaningful information with streamflow or discharge necessary to quantify water availability and hydrological processes.

\begin{figure}
\centering
    \includegraphics[width=0.5\textwidth]{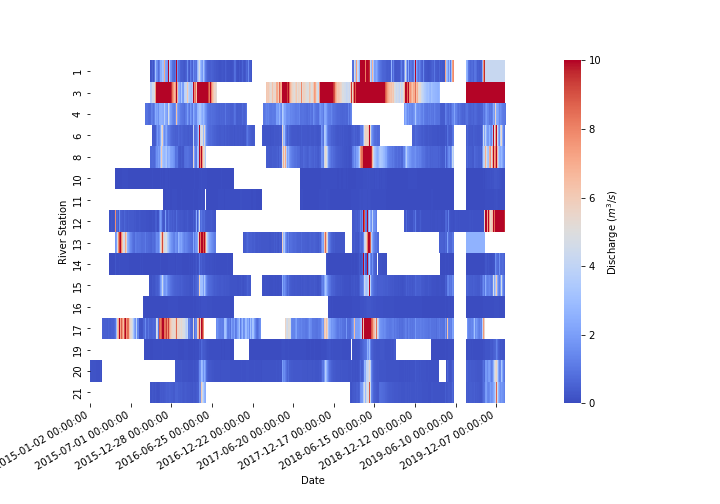}
    \caption{Visualisation of range and availability of streamflow measurements ($\mathrm{m^3}/s$) at each of the 19 stations. The colorbar indicates values of streamflow while whitespace denotes periods where no data were collected.}
    \label{fig:observations}
\end{figure}

\pagebreak
Water level data were converted to streamflow using established heuristic relationship between water level and discharge (termed stage\textendash discharge rating curve analysis) \cite{turnipseed2010discharge}. This required the collection of streamflow data, development of water level-streamflow rating curve or relationships, and using the generated relationship to convert water level measurements to streamflow. Flow measurements were collected  using  an  Acoustic  Doppler  Current  Profiler  (ADCP)  instrument.   An  ADCP  is  a  type  of sonar  that  measures  and  records  water  current  velocities  over  a  range  of  depths using the Doppler effect of sound waves scattered back from particles within the water column. Point measurements were collected at multiple locations across the river channel and these used to compute the streamflow discharge at a point in time. Collecting these measurements under multiple streamflow conditions allows one to create the water level discharge relationship. In our case, flow measurements were collected at each station between 50\textendash200 times over the study period.
Where sufficient flow data were not available robust streamflow relationships could not be generated. 
The HOBO water level measurements were converted to streamflow measurements using the above process (19 of the 26 stations had suitable flow measurements) and data were resampled from 30 minute intervals to daily values.
Figure \ref{fig:observations} presents a heatmap illustrating range of observed values and also highlighting challenges related to data sparsity. For most stations, there are periods when data was not recorded, extending for up to 50\% of the time at some stations. While this is a challenge for DL, it also highlights the possibility of using DL models to address data gaps and inform hydrology processes. 

\vspace{10mm}

\begin{figure*}[ht]
  \centering
  \includegraphics[width=0.9\linewidth]{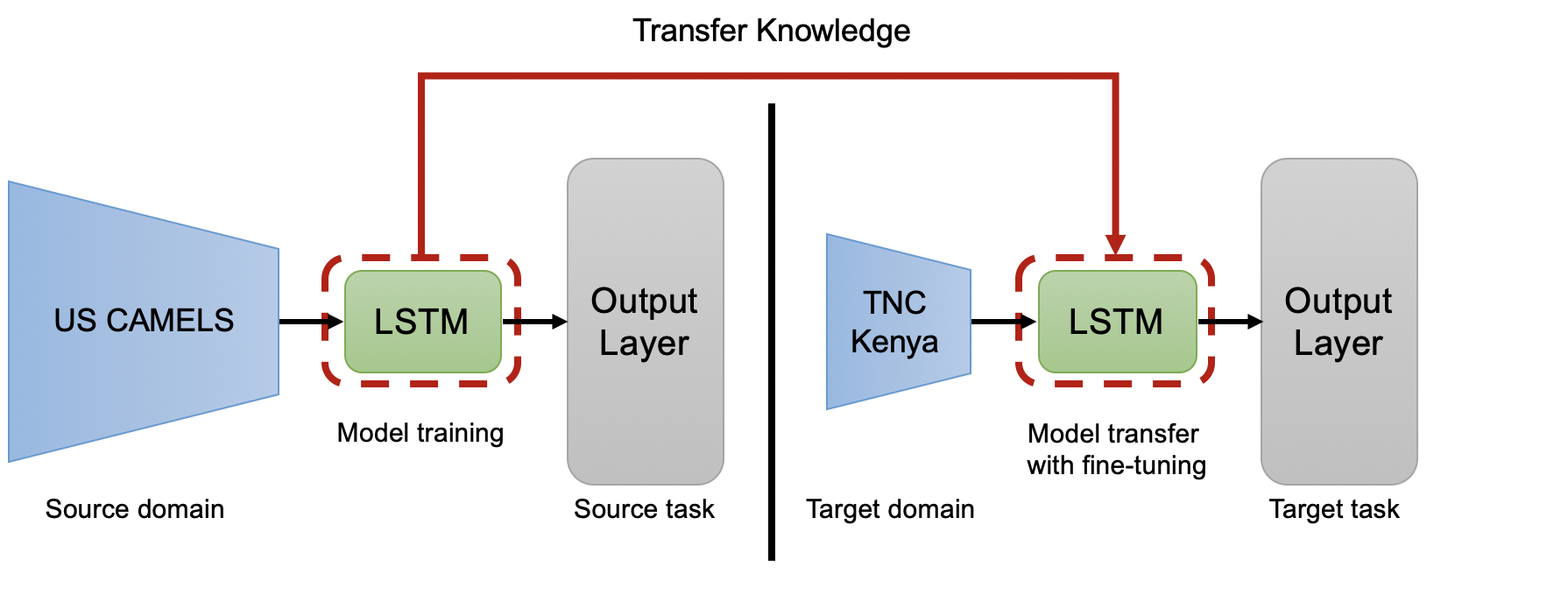}
  	  \caption{Main architecture of the transfer learning approach in which the knowledge learned from training over a fairly large source domain dataset (US CAMELS) is applied to a smaller and sparse target dataset (Kenya) for generalization performance of streamflow prediction.}
  \label{fig:lstm-main-arch}
\end{figure*}

\section{Proposed Approach}
In this section, we first introduce a standard recurrent neural network, specifically Long Short-Term Memory (LSTM), that uses a loss function commonly used in hydrology. We then describe in detail the method behind our transfer learning approach.

\subsection{Model Architecture}
We propose a multi-layer LSTM network  as the base model to perform transfer learning through two separate domains, as shown in Figure \ref{fig:lstm-main-arch}. We set up our deep learning architecture from the open-source work of \cite{kratzert2019toward,kratzert2019towards}, where the inputs are dynamic features from water catchment processes such as: min/max air temperature, vapor pressure, and precipitation. The aim of the hydrological LSTM model is to learn the prediction of the continuous streamflow response on a target dataset (Kenya) from the long-term dependent features given the knowledge learned from a large source domain dataset (US CAMELS). 


To calculate the model's performance of streamflow prediction, we present the use of Nash-Sutcliffe efficiency (NSE) \cite{nash1970river}, which is a statistical estimator that measures the predictive skill of a hydrological model. NSE is defined as:


\begin{equation}
NSE = 1 - \frac{ \sum_{t=1}^{T} (Q_{m}^{t} - Q_{o}^{t})^2}{ \sum_{t=1}^{T} (Q_{o}^{t} - \bar{Q}_{o})^2},
\label{eq1}
\end{equation}

where \(Q_{m}^{t}\) is the modeled discharge (streamflow) at time \(t\) , and \(Q_{o}^{t}\) is the observed discharge at time \(t\). The value of the NSE varies on the interval \( (-\infty, 1] \), where a set of conditions are used as criteria for understanding model performance. \( NSE = 0 \) denotes equivalent predictive skill of the model to the mean of the time-series (i.e., sum of squared errors). In a perfect model situation, \( NSE = 1 \) is when the estimation error variance equal to zero. If \( NSE \leq 0 \), the estimation error variance of the model is larger than the variance of observation. Hence, if the observed mean is a better predictor than the model, then the model has little predictive skill.

Similar models such as mean squared error (MSE) are widely adopted for calibration and model evaluation with observed data \cite{gupta2009decomposition}. Unlike MSE, NSE does not overweight catchments with larger streamflow values, which could have influence on the models performance \cite{kratzert2019towards} (Figure \ref{fig:observations} indicates the large data ranges involved). Hence, this study focuses solely on the performance of the hydrological model through the NSE coefficient.

\subsection{Transfer Learning Approach}
Given a LSTM model \( f \) that fits over a domain dataset \( D = {(X, Y)} \) where \( X \) is the set of inputs \( (x_{1}, x_{2}, ..., x_{n}) \) and \( Y  \)  is the set of labels \( (y_{1}, y_{2}, ..., y_{n}) \), we define a source source domain dataset to be \( D_{s} = {\{(x_{s}^{i}, y_{s}^{i})\}}_{i=1}^{m_{s}} \) and a target domain dataset to be \( D_{t} = {\{(x_{t}^{i}, y_{t}^{i})\}}_{i=1}^{m_{t}} \). We also define a domain task \( T =  {(Y, f(X))}\) where we denote the source task domain as \( T_{s} = \{ {(y_{s}^{i}, f(x_{s}^{i}))} \} \) and the target task domain as \( T_{t} = \{ {(y_{t}^{i}, f(x_{t}^{i})) } \} \). The objective transfer function learning aims to learn the target conditional probability distribution \( P = (Y_{t} | X_{t}) \) in \( D_{t} \) with the knowledge gained from both \( D_{s} \) and \( T_{s} \), where \( D_{s} \neq D_{t} \) or \( T_{s} \neq T_{t} \) \cite{pan2009survey}. 

The source and target domains share similar feature input space, however labels \( Y_{s} \) and \( Y_{t} \) are heterogeneous. Hence, the LSTM model \( f \) cannot be fine-tuned directly on the target domain dataset \( D_{t} \) due to the discrepancy of task objectives. A common technique to transfer knowledge and fine-tune is to split \( f \) between an LSTM model representation \( F_{\bar{\theta}} \) (parametrized by \( \bar{\theta} \)) and the task-oriented head–which is the topmost layer used for regression–denoted as \( G_{\theta_{s}} \) (parametrized by \( \theta_{s} \)). The head of the LSTM model can be replaced with \( H_{\theta_{t}} \) (parametrized to \( \theta_{t} \)), where a random initialization is used to fit the distribution of target domain label \( Y_{t} \). We can then perform a standard fine-tuning method over a loss function in Equation \ref{eq2} for parameter optimization: 
\begin{equation}
    \argmin_{\bar{\theta}, \theta_{t}} \frac{1}{|D_{t}|} \sum_{i=1}^{m_{t}} L(H_{\theta_{t}}(F_{\bar{\theta}}(x_{t}^{i}, y_{t}^{i}))).
    \label{eq2}
\end{equation}

\noindent Using Equation \ref{eq1}, we apply the replacement and initialization of the LSTM regression layer to refine the NSE loss function as:

\begin{equation}
    \argmin_{\bar{\theta}, \theta_{t}}~ 1 - \frac{ \sum_{t=1}^{m_{t}} (H_{\theta_{t}}(F_{\bar{\theta}}(x_{t}^{i})) - y_{t}^{i})^{2} }{ \sum_{t=1}^{m_{t}} (y_{t}^{i} - \bar{ y}_{t}^{i})^2}.
    \label{eq3}
\end{equation}

\noindent The loss function in Equation \ref{eq3} enables the LSTM model to use its learned weights as a starting point for training as opposed to traditional random initialization methods such as Gaussian distribution. Hence, this yields an opportunity at improving the generalization performance of predicting streamflow response. 

\begin{table*}[ht]
\centering
\begin{tabular}{l|c c c c c c}
    \hline
    \multicolumn{7}{c}{TNC Dataset in Kenya Region} \\
    \hline
    Model & $\mathrm{Median_{NSE}}$ & $\mathrm{Mean_{NSE}}$ & $\mathrm{Max_{NSE}}$ & $\mathrm{Min_{NSE}}$ & $\mathrm{STD}$ & $\mathrm{NSE > 0}$ \\
    \hline
    $\mathrm{LSTM}$ & -0.85 & -30.70 & 0.58 & -1259.73 & 2.08 & 2 \\
    \hline
    $\mathrm{LSTM_{SCA}}$ & -0.93 & -19.70 & 0.62 & -423.45 & 1.29 & 1 \\
    \hline
    $\mathrm{LSTM_{TL}}$ & \textbf{-0.49} & \textbf{-14.33} & 0.53 & \textbf{-314.46} & \textbf{0.46} & 3 \\
    \hline
    $\mathrm{LSTM_{TL + SCA}}$ & -0.51 & -17.48 & \textbf{0.67} & -577.23 & 1.02 & 3 \\
    \hline 
\end{tabular}
\caption{Model comparison of the proposed methods using the NSE model estimation. The statistics were averaged across 30 runs using a different seed each time.}
\label{tab: table1}
\end{table*}

\section{Experiments}
\subsection{Experimental Setup}
\textbf{Datasets}~ We collected data from two primary sources: (i) US CAMELS and (ii) TNC. The US CAMELS \cite{newman2014development} is a large sample benchmark dataset that has covered temporal and geospatial information related hydrology across 531 basins from October 1st, 1980 through September 30th, 2014. All of the LSTM model calibration – which is utilized for transfer learning – was performed from October 1st, 1999 through September 30th, 2008, while the evaluations were performed from October 1st, 1989 through September 30th, 1999. The total time period used for model simulation over the geospatial input data amounts to 10,227 days (28 years).

The Kenya dataset contained hydrological data from 2015-2019, in which 26 river stations were gauged. However, only 12 had sufficient hydrological data (i.e. observations were available $\mathrm{>50\%}$ of time) to use for this study. We combined with geospatial static catchment attributes to characterise the hydrological profile of the area similar to the US CAMELS data. These categorical static attributes included: catchment area, elevation, and slope, mean precipitation, high and low precipitation frequency and duration, mean evapotranspiration, and aridity (ratio of precipitation and evapotranspiration), and vegetation indices such as leaf area index and NDVI.  
Suitably conditioned data were available from  01/09/2015\textendash16/10/2019 which was split into training between period 01/09/2015\textendash22/02/2018 and testing from 22/02/2018\textendash16/10/2019.
The total time period used for this experiment is 1,507 days ( $>$ 4 years). In both datasets, we leverage the same subset of geospatial features which include land cover, climate, soils, as well as dynamic meteorological descriptors. The target variable is a continuous value of the daily observed streamflow (or discharge) at each station. \\

\noindent \textbf{Model Comparison}~ We present the following models used for this experiment:
\begin{itemize}
    \item \textit{Standard LSTM}: We employ a standard multi-layered LSTM network based on the existing work on predicting streamflow~\cite{kratzert2019toward,kratzert2019towards}. The output layer uses a regression head to simulate the discharge over unseen data.
    \item \textit{LSTM with static catchment attributes(SCA)} ($\mathrm{LSTM_{SCA}}$): We incorporate static basin descriptors related to land cover, soil, climate, vegetation, and topography that has shown to augment ability for the LSTM to generalize~\cite{kratzert2019toward}. 
    \item \textit{LSTM with transfer learning} ($\mathrm{LSTM_{TL}}$): We present the existing transfer learning solution using an LSTM over the US CAMELS dataset. The weights are transferred to generalize over the Kenya dataset.
    \item \textit{LSTM with TL + SCA} ($\mathrm{LSTM_{TL + SCA}}$): We combine the ability of additional catchment attributes to augment the model performance along with our proposed transfer learning approach. 
\end{itemize}

\noindent \textbf{Hyperparameters and Evaluation Metrics}~ The hyperparameters of the baseline models were employed uniformly. The LSTM network structure consisted of 128 hidden units and 1 output regression layer used for forecasting daily streamflow over 30 epochs. We initialize the learning rate \( \eta \) to 0.001 during the first epoch and adjust to 0.0005 over the remaining epochs. To penalize inconsistent predictions, we implement an ADAM optimizer as our regularization term.

Our primary evaluation metric for this study is the Nash-Sutcliffe model Efficiency estimation (NSE) \cite{nash1970river}, which estimation of predicting discharge compared to similar metrics such as MSE and RMSE. We leveraged six statistical metrics from our generated NSE values to calculate the performance of the propose models which include: median ($\mathrm{Median_{NSE}}$), mean ($\mathrm{Mean_{NSE}}$), maximum ($\mathrm{Mean_{NSE}}$), min ($\mathrm{Min_{NSE}}$), standard deviation ($\mathrm{STD}$), and number of basins where the NSE value is greater than 0 ($\mathrm{NSE>0}$).


\begin{figure}
\centering
    \includegraphics[width=0.5\textwidth]{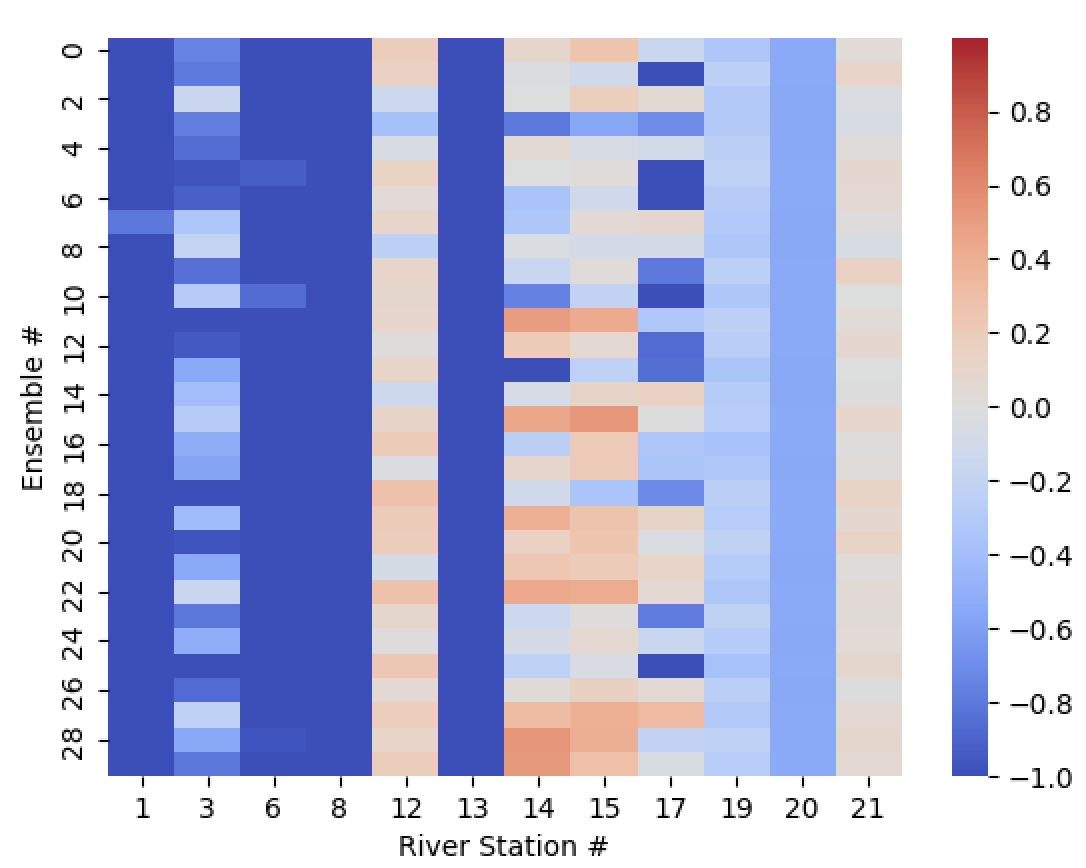}
    \caption{A colormap visualization that demonstrates the performance of the fine-tuned LSTM$_{TL}$ model and its response to generalize on sparse data inputs over 30 runs.}
    \label{fig:lstm_colormap}
\end{figure}

\subsection{Comparing Proposed Methods}
We evaluated the performance of each proposed model over the TNC Kenya dataset and compared their results based on our aforementioned metrics. From Table \ref{tab: table1}, the $\mathrm{LSTM_{TL}}$ displayed the best overall performance, notably achieving a higher $\mathrm{Median_{NSE}}$ and $\mathrm{Mean_{NSE}}$ score. The optimized weights served as a better starting point for the target LSTM model than random initialization. Augmenting the standard LSTM model with static catchment attributes (i.e., $\mathrm{LSTM_{SCA}}$,  $\mathrm{LSTM_{TL+SCA}}$) showed a lack of improvement in performance relative to their counterparts. While some modest improvement of mean NSE and standard deviation were reported by addition of SCA to baseline LSTM, this did not translate to the model with transfer learning. This likely results from the different scale of geospatial datasets used in the CAMELS and Kenya dataset which may require some statistical conditioning to resolve more closely. As an example climate indicators for Kenya were extracted from the ECMWF ERA5 global reanalysis product at horizontal resolution of $\sim$\SI{28}{km} while the US CAMELS dataset used the US National N15 Daymet dataset at resolution of $\sim$\SI{1}{km}.  The difference in the standard deviation (Std) between the standard LSTM model and $\mathrm{LSTM_{TL}}$ (2.08 vs. 0.46) was largely due to the lack of temporal streamflow observations that cause large gaps during model training and validation. Employing transfer learning from a model train on a significantly richer source domain enables the model to reduce deviation for producing more effective results on unseen streamflow.

Figure \ref{fig:lstm_colormap} reports the performance of 30 $\mathrm{LSTM_{TL}}$ model instances trained with random initialization. It indicates that 4 of the 12 stations reported $\mathrm{NSE<<0}$ while 3 reported $\mathrm{NSE>0}$ and further 3 had values $\sim0$. While these values return moderate predictive skill, they present a promising demonstration of the potential of transfer learning to address the combined challenge of data sparsity, data uncertainty, and complex physical and empirical relationships.

\begin{figure}[t!]
\centering
    \includegraphics[width=0.5\textwidth]{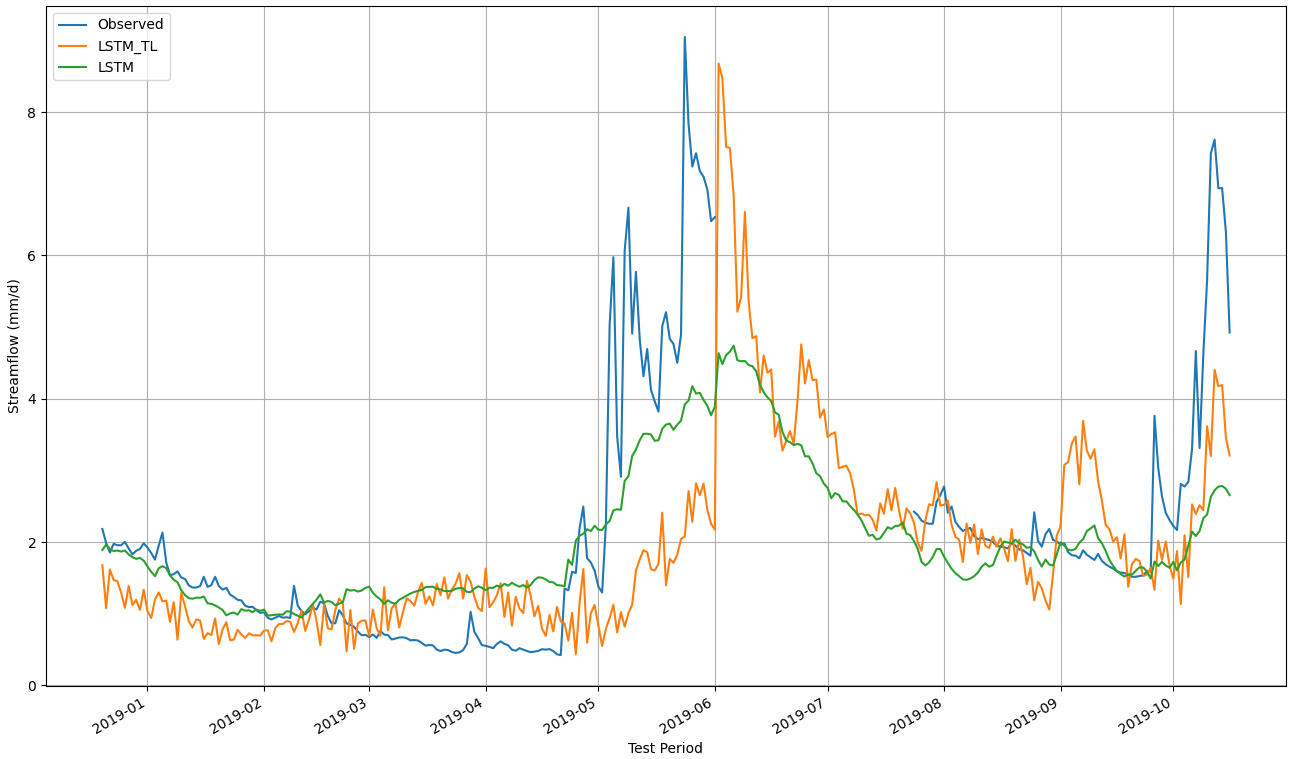}
    \caption{Comparisons on the prediction of streamflow at a given basin between the baseline LSTM and LSTM$_{TL}$ model.}
    \label{fig:pred_streamflow}
\end{figure}

In Figure \ref{fig:pred_streamflow}, we compare the performance of the proposed transfer learning model ($\mathrm{LSTM_{TL}}$) with a baseline model to predict the streamflow response at a given station. Herein, demonstrates the ability of the transfer learning model to address the limitations of data collection in areas where geospatial
information is sparse and inconsistent. Compared to the standard LSTM model, ($\mathrm{LSTM_{TL}}$) provides a more realistic prediction over the TNC Kenya dataset by leveraging the learned weights from a data-rich source domain dataset (US CAMELS). Thus, the simulated streamflow  from the ($\mathrm{LSTM_{TL}}$) model can be of more effective use in informing stakeholders in developing areas (e.g., farmers, Kenyan water funds) about the water availability within the Kenyan region for better water resource management.

\section{Discussion}
While employing transfer learning has many implications of success, as demonstrated by the improved performance of the standard LSTM model, there still presents many concerns regarding its applicability for time series tasks. Successful transfer learning theories and application has largely been applied to computer vision tasks \cite{long2017deep,zamir2018taskonomy}. Transfer learning has also seen success in multi-lingual machine translation as it has the ability to create word representation (embeddings) \cite{zoph2016transfer,aji2020neural}. However, this success has not been extended to time-series related tasks due to the lack of existing literature in both theory and application. Time-series models may be elusive, as it is harder to find a useful hierarchy or a set of intermediate representations that could generalize to different domains. In addition, negative transfer learning is a common challenge as it can negatively affect performance when performing knowledge transferring from a source to a target domain \cite{torrey2010transfer}.

Furthermore, the initiatives to improve hydrological modeling have equally, if not more so, relied on sufficient data collection processes as it does on the model's underlying algorithm. The sparsity and inconsistency of the Upper Tana dataset caused performance issues related to the predictive streamflow skill. This can be due to the lack of equipment used to track meteorological data such as precipitation and air temperature in developing countries. The US and many areas across Europe, on the other hand, have had the means of curating large-sample datasets through the accessible resources of weather tracking at their disposal. The insufficient data of tracking meteorological information in countries from Africa presents an issue of informing stakeholders (e.g., farmers, water funds) about the water quality and availability for proper water resource management. Programs such as The trans-African hydro-meteorological observatory (TAHMO) \cite{van2014trans}  which aims for continuous monitoring of precipitation data across Sub-Saharan Africa are key initiatives to allow for more effective, applied DL to hydrology. In addition, the resolution provided by different geoscientific monitoring and modelling tools such as NASA and European Centre for Medium Weather Forecasting (ECMWF) at continental and global scales could be reconciled towards higher resolution in places such as Kenya (to provide resolution comparable to that currently provided for US and Europe by the respective organisations). Approaches for statistical downscaling of these global datasets (e.g. \cite{jakob2011empirical}) could improve the performance of transfer learning across different geographies. 

\section{Conclusion}
The science of hydrology has largely derived from using known physical and empirical relationships -- guided by user expertise and intuition --  to resolve complex rainfall --  discharge relationships. Huge potential exists to use deep learning to augment these inference and intuition from large hydrological datasets. A fundamental challenge currently is data sparsity. 

In this paper, we address the limitations of streamflow predictions in data sparse regions using transfer learning. 
We first address the underlying problem by collecting, pre-processing, and curating a dataset, (TNC Kenya), similar to that of US CAMELS, which has the same dynamic meteorological input features and the same set of static features describing each catchment area. Our approach attempts to generalize a model in small target domain datasets given knowledge learned from a significantly larger source domain dataset by knowledge transfer. We employ a multi-layer LSTM model to forecast streamflow response using the dynamic meteorological inputs as well as static catchment descriptors. We experimented with the predictive skill of forecasting streamflow under four conditions concerning transfer learning, static catchment attributes, both, or neither. Our results demonstrate the improved performance and potential of transfer learning to give insights on streamflow prediction in areas where data is sparse due to the lack of temporal and geospatial resources. Our future work aims to look at other drivers of predictive streamflow forecasting using spatial-temporal methods such as graph neural networks and convolutional neural networks.

\bibliographystyle{aaai} 
\bibliography{main}

\end{document}